\documentclass[10pt,twocolumn,letterpaper]{article}

\usepackage{cvpr}              % To produce the CAMERA-READY version
\usepackage{graphicx}
\usepackage{amsmath}
\usepackage{amssymb}
\usepackage{booktabs}

\usepackage[super]{nth}

\usepackage{fixltx2e}

\usepackage{caption}
\usepackage{subcaption}
\usepackage[pagebackref,breaklinks,colorlinks]{hyperref}

\usepackage[capitalize]{cleveref}
\crefname{section}{Sec.}{Secs.}
\Crefname{section}{Section}{Sections}
\Crefname{table}{Table}{Tables}
\crefname{table}{Tab.}{Tabs.}

\def\cvprPaperID{9} % *** Enter the CVPR Paper ID here
\def\confName{CVPR}
\def\confYear{2023}

\begin{document}

\title{Global Motion Understanding in Large-Scale Video Object Segmentation}

\author{Volodymyr Fedynyak\\
Ukrainian Catholic University\\
{\tt\small v.fedynyak@ucu.edu.ua}
\and
Yaroslav Romanus\\
Ukrainian Catholic University\\
{\tt\small yaroslav.romanus@ucu.edu.ua}
\and
Oles Dobosevych\\
Ukrainian Catholic University\\
{\tt\small dobosevych@ucu.edu.ua}
\and
Igor Babin\\
ADVA Soft\\
{\tt\small ihor.babin@adva-soft.com}
\and
Roman Riazantsev\\
ADVA Soft\\
{\tt\small roman.riazantsev@adva-soft.com}
}
\maketitle

%%%%%%%%% ABSTRACT
\begin{abstract}
    In this paper, we show that transferring knowledge from other domains of video understanding combined with large-scale learning can improve robustness of Video Object Segmentation (VOS) under complex circumstances. Namely, we focus on integrating scene global motion knowledge to improve large-scale semi-supervised Video Object Segmentation. Prior works on VOS mostly rely on direct comparison of semantic and contextual features to perform dense matching between current and past frames, passing over actual motion structure. On the other hand, Optical Flow Estimation task aims to approximate the scene motion field, exposing global motion patterns which are typically undiscoverable during all pairs similarity search. We present WarpFormer, an architecture for semi-supervised Video Object Segmentation that exploits existing knowledge in motion understanding to conduct smoother propagation and more accurate matching. Our framework employs a generic pretrained Optical Flow Estimation network whose prediction is used to warp both past frames and instance segmentation masks to the current frame domain. Consequently, warped segmentation masks are refined and fused together aiming to inpaint occluded regions and eliminate artifacts caused by flow field imperfects. Additionally, we employ novel large-scale MOSE 2023 dataset to train model on various complex scenarios. Our method demonstrates strong performance on DAVIS 2016/2017 validation (93.0\% and 85.9\%), DAVIS 2017 test-dev (80.6\%) and YouTube-VOS 2019 validation (83.8\%) that is competitive with alternative state-of-the-art methods while using much simpler memory mechanism and instance understanding logic.
\end{abstract}

%%%%%%%%% BODY TEXT
\section{Introduction}
\label{sec:intro}

Video Object Segmentation (VOS) is a fundamental task in Video Understanding, aiming to segment multiple objects through an entire video sequence. In this work, we address semi-supervised video object segmentation, i.e. the scenario where only the first frame annotations are given, or the annotations are given only for the frames where the corresponding object appears in the video for the first time.

The key feature of Video Object Segmentation is the complete agnosticity of the actual class information for considered objects. This allows a very broad range of possible applications, including but not limited to autonomous driving, sports and video editing.

Prior works achieved significant success in VOS, focusing on making solution highly generalizable and robust under different complex scenarios while maintaining real-time efficiency and low GPU memory footprint. AOT \cite{yang2021associating} proposed to map objects to a pre-defined set of feature vectors making possible simultaneous processing of many instances. While most works use feature memory to correctly treat occlusions and eliminate errors during propagation, XMem \cite{cheng2022xmem} points out the high memory consumption of such an approach and designs efficient unified multi-type memory inspired by Atkinson-Shiffrin model. DeAOT \cite{yang2021decoupling} notes the poor performance of existing methods when the objects drastically change in scale and appearance during the video, presenting a novel feature decoupling block to treat such cases more robustly. ISVOS \cite{wang2022look} argues that instance understanding matters in VOS and employ an instance segmentation branch based on state-of-the-art instance segmentation architectures increasing the VOS performance for video clips with a high number of similar objects.

Existing approaches rely on dense attention-based feature matching \cite{vaswani2017attention} to propagate segmentation masks through the video sequence. Even though this achieves remarkably high scores on existing benchmarks, a single all-pairs correlation search is not capable of capturing global motion context and uncovering relevant motion patterns. In this work, we argue that motion understanding matters in VOS. Inspired by ISVOS proposing to reuse existing instance segmentation architectures to improve instance understanding for VOS domain, we propose to reuse existing optical flow estimation architectures to propagate instance information between video frames.

We present WarpFormer, an VOS architecture that benefits from global motion structure knowledge. We adopt a generic VOS architecture for spatial-temporal matching similar to \cite{yang2021associating} and replace short-term memory mechanism with optical flow warp, for which we employ a flow estimation network. The propagation process is tackled by optical flow warp while the spatial windowed attention is used to refine warped segmentation mask and inpaint occlusions. Finally, refined mask is fused with long-term memory matches and passed to decoder.

We conduct additional training of our model on large-scale MOSE 2023 \cite{MOSE} dataset to achieve robustness under complex VOS scenarios. We evaluate our method on DAVIS 2016 \& 2017 and YouTube-VOS 2019 benchmarks. Conducted experiments demonstrate that both exploiting global motion structure and large-scale training improve evaluation scores and qualitative results. 

%------------------------------------------------------------------------
\section{Related Work}
\label{sec:formatting}

\begin{figure*}[ht]
    \centering
  \includegraphics[width=0.9\linewidth]{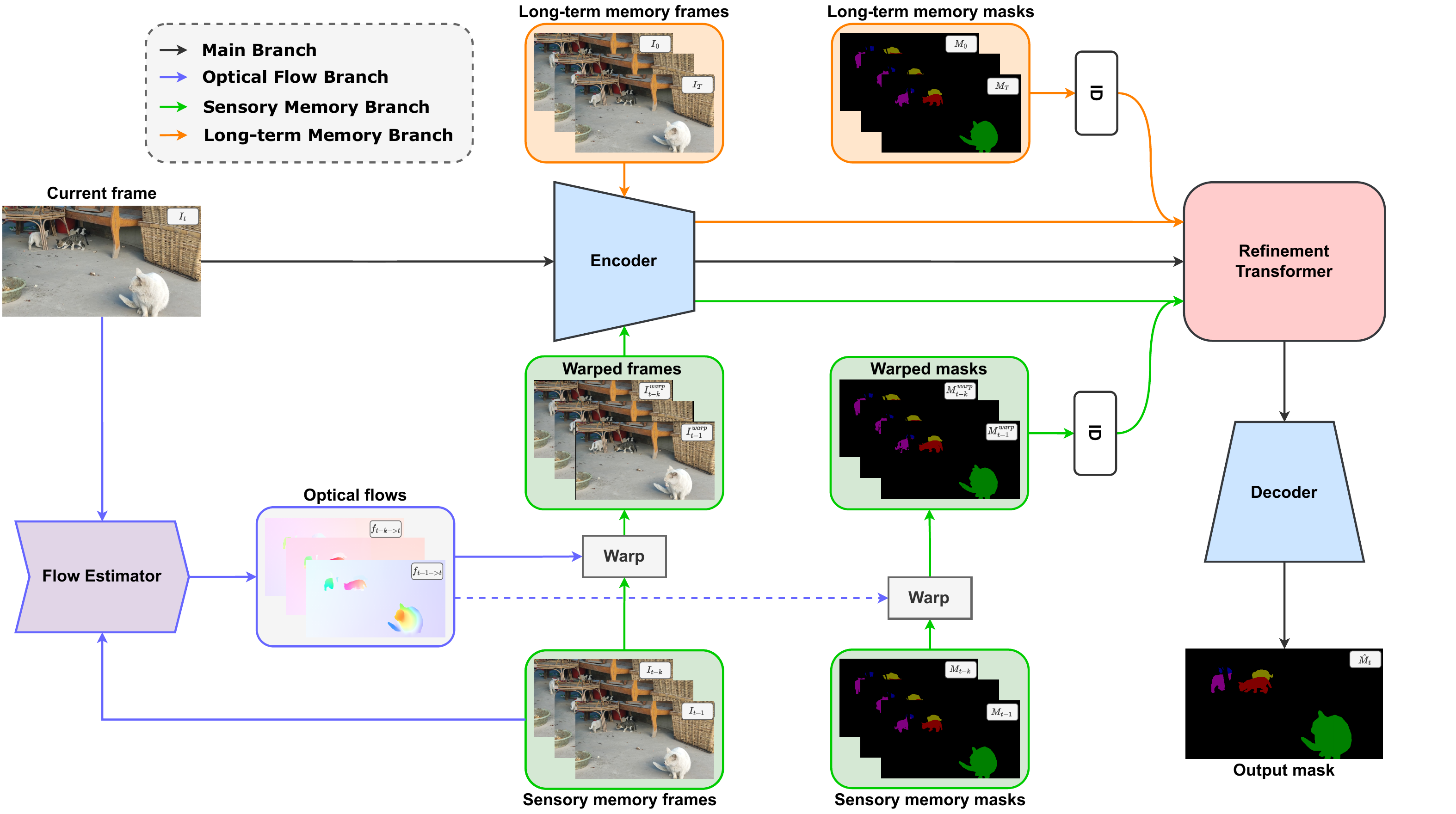}
  \caption{\textbf{The architecture of a WarpFormer}. Best viewed in color.}
  \label{fig:warpformer}
\end{figure*}

\subsection{Optical Flow Estimation}

Optical flow is a critical component of our work. Its main idea is to estimate the shift of all points from one frame to another. Early approaches in that area were focused on optimization problems, maximizing visual similarity with regularization terms \cite{horn1981determining, black1993framework, bruhn2005lucas, sun2014quantitative}. However, the advent of deep neural networks, specifically convolutional networks, propelled the field forward. Pioneering models like FlowNet \cite{dosovitskiy2015flownet} and FlowNet2.0 \cite{ilg2017flownet} set the stage for more advanced methods, such as SpyNet \cite{ranjan2017optical}, PWC-Net \cite{sun2018pwc}, and LiteFlowNet \cite{hui2018liteflownet} which adopted coarse-to-fine and iterative estimation strategies.

Despite their advancements, these models struggled to capture small, fast-moving objects during the coarse stage. The RAFT model \cite{teed2020raft} introduced significant improvements, a novel architecture employing a coarse-and-fine (multi-scale search window per iteration), and a recurrent approach to optical flow estimation. Subsequent works based on RAFT, such as GMA \cite{jiang2021learning} and DEQ-Flow \cite{bai2022deep}, aimed to improve computational efficiency or enhance flow accuracy.

A recent example of a state-of-the-art recurrent approach is FlowFormer \cite{huang2022flowformer}---an extension of the RAFT architecture. It introduces a transformer-based method that aggregates cost volume in a latent space. This approach builds on the work of Perceiver IO \cite{jaegle2021perceiver}, which was the first to incorporate transformers \cite{vaswani2017attention} for establishing long-range relationships in optical flow, achieving state-of-the-art performance. FlowFormer retains the cost volume as a compact similarity representation and pushes the search space to the extreme by globally aggregating similarity information using a transformer architecture. Another state-of-the-art approach is GMFlow \cite{xu2021gmflow}, which formulates optical flow as a global matching problem and employs a customized Transformer for feature enhancement, global feature matching, and flow propagation. This approach outperforms the RAFT on the Sintel benchmark while offering greater efficiency \cite{xu2021gmflow}.

% \subsection{Attention-based Image matching}
% TODO: do we need this section?

\subsection{Video Object Segmentation}

One popular method that has achieved state-of-the-art performance in VOS is AOT (Associating Objects with Transformers for VOS) \cite{yang2021associating}. AOT exploits the Long Short-Term Transformer (LSTT) block that includes self-attention, short-term attention, and long-term attention to extract features from input images. Long-term attention is responsible for aggregating information from long-term memory frames, while short-term attention propagates information from the previous frame. The outputs of long-term and short-term attention blocks are combined in the feed-forward network, which passes information to the decoder that returns mask estimation for the current frame. AOT also uses a joint architecture that includes an attention map for the attention blocks and a 4D correlation volume, as in the RAFT \cite{teed2020raft} architecture, to calculate the same spatial correlation between frames. The short-term attention in AOT and 4D correlation volume in RAFT calculate the same correlation between features from consecutive frames, which can be combined in the shared part of the joint architecture as a unified motion representation.

 DeAOT \cite{yang2021decoupling} (Decoupling Features in Hierarchical Propagation for Video Object Segmentation) is a recent method for semi-supervised video object segmentation that builds on the hierarchical propagation introduced in the AOT approach. DeAOT decouples the hierarchical propagation of object-agnostic and object-specific embeddings into two independent branches to prevent the loss of object-agnostic visual information in deep propagation layers. To compensate for the additional computation from dual-branch propagation, DeAOT introduces a Gated Propagation Module that is carefully designed with single-head attention. Experimental results show that DeAOT outperforms AOT in both accuracy and efficiency, achieving new state-of-the-art performance on several benchmarks, including YouTube-VOS, DAVIS 2016 and DAVIS 2017.

\subsection{Optical Flow-based Video Segmentaiton}

Optical flow-based Video Object Segmentation has progressed substantially over time. One of the early works in this domain, MaskTrack~\cite{khoreva2016learning}, combined object segmentation and tracking by employing optical flow for object mask propagation and refining the results using a convolutional neural network (CNN). Building on this foundation, OSVOS~\cite{caelles2017one} further enhanced segmentation performance.

More advanced methods like PDB~\cite{zhang2019pdb} and RVOS~\cite{ventura2019rvos} emerged, employing multi-stage frameworks and recurrent neural networks, respectively, while still leveraging optical flow. The Regional Memory Network (RMNet)~\cite{xie2021rmnet} recently introduced a local-to-local matching approach that minimizes mismatches with similar objects using regional memory embedding and optical flow-based tracking.

We build on these foundational works in our proposed method, utilizing optical flow for short-term frames and attention mechanism for long-term frames to enhance the segmentation process.

\section{Method}

\subsection{Background}

Video object segmentation is a challenging task that often involves tracking multiple objects of interest in a single video. Previous approaches to this problem have focused on matching and propagating a single object, requiring independent matching and propagation of each object in multi-object scenarios \cite{voigtlaender2019feelvos}. This can result in increased GPU usage and inference time, hindering the efficiency of the overall pipeline.

To address this challenge, AOT proposed an identification mechanism for embedding masks of any number into the same high-dimensional space, enabling multi-object scenario training and inference as efficient as single-object ones \cite{yang2021associating}. This mechanism involves creating a predefined set of $M$ trainable vectors, known as the identity bank, and picking a vector from this bank for each pixel corresponding to a specific class. During training, the vector corresponding to each class is randomly selected to ensure uniform training of the identity bank.
To add object-specific information to the feature maps in our architecture, which are at the $\frac{1}{16}$ spatial size of the input video, we adopt a patch-wise identity bank strategy similar to AOT~\cite{yang2021associating}. This involves dividing the input mask into non-overlapping $16\times16$ patches, matching each pixel in the patch with the corresponding vector from the identity bank, and obtaining the final result for the identity bank by summing the values for the pixels inside the patch. This operation also encodes some geometry inside the patch and can be implemented as a single $16\times16$ convolution.

\subsection{Warp Refinement Transformer}

\begin{figure*}[ht]
\caption{\textbf{Warpformer Modules}. Best viewed in color.}
    \begin{subfigure}[c]{0.3\linewidth}
    \centering
    \includegraphics[width=\linewidth]{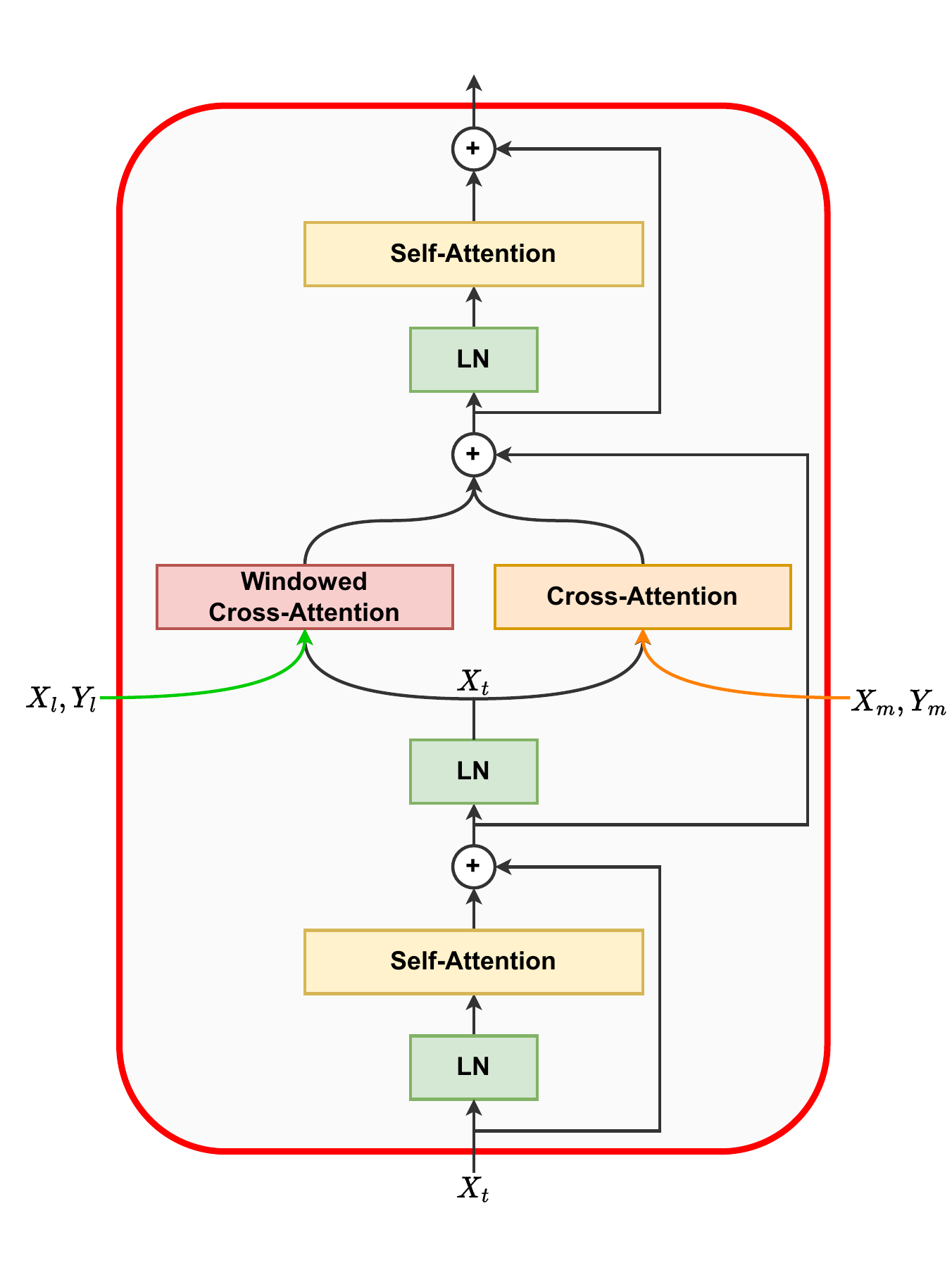}
    \caption{Refinement Transformer Block}
    \label{fig:refinement}
    \end{subfigure}\hfill
    \begin{subfigure}[c]{0.3\linewidth}
    \centering
    \includegraphics[width=\linewidth]{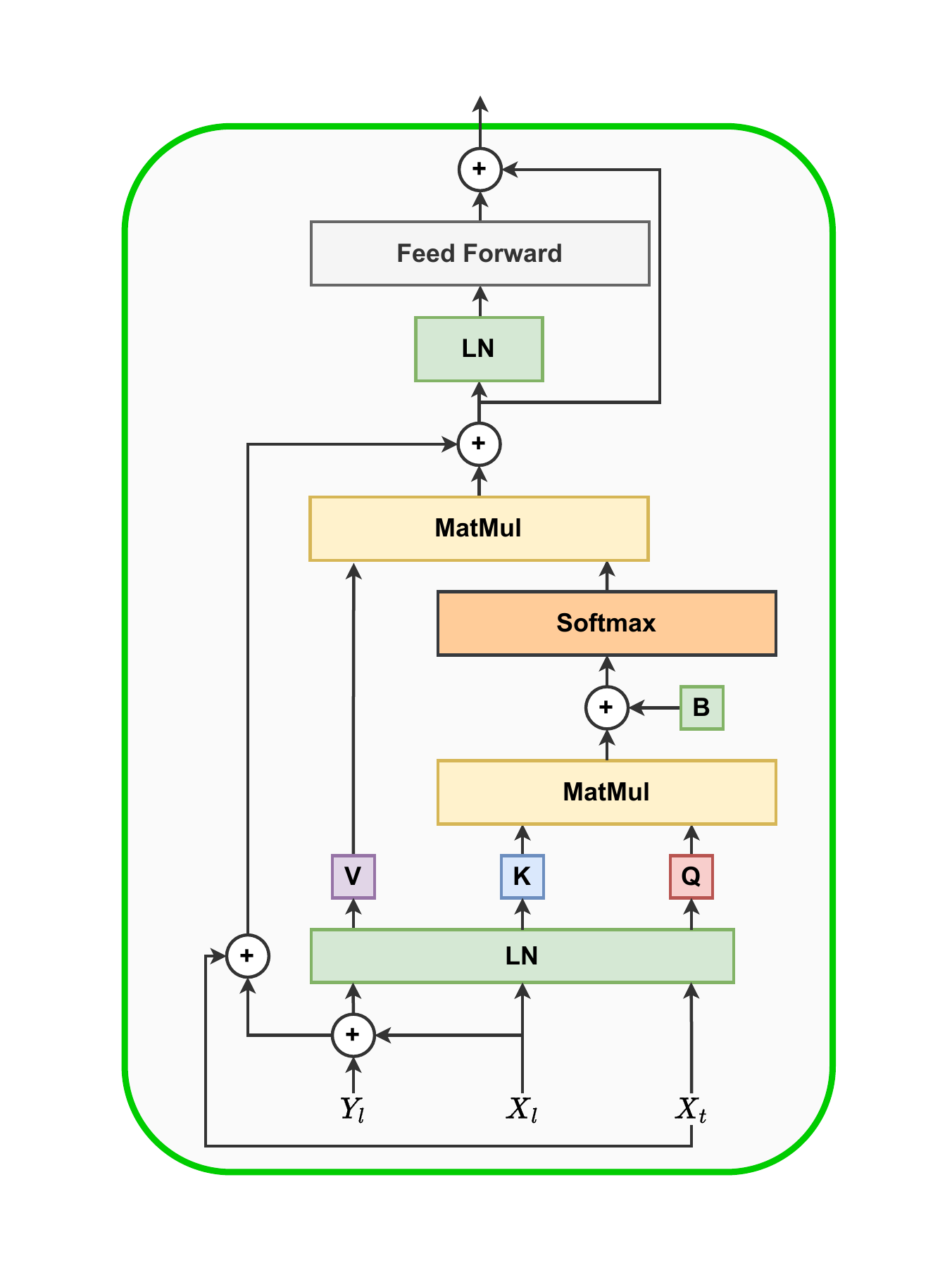}
    \caption{Windowed Cross-Attention}
    \label{fig:short}
    \end{subfigure}\hfill
    \begin{subfigure}[c]{0.3\linewidth}
    \centering
    \includegraphics[width=\linewidth]{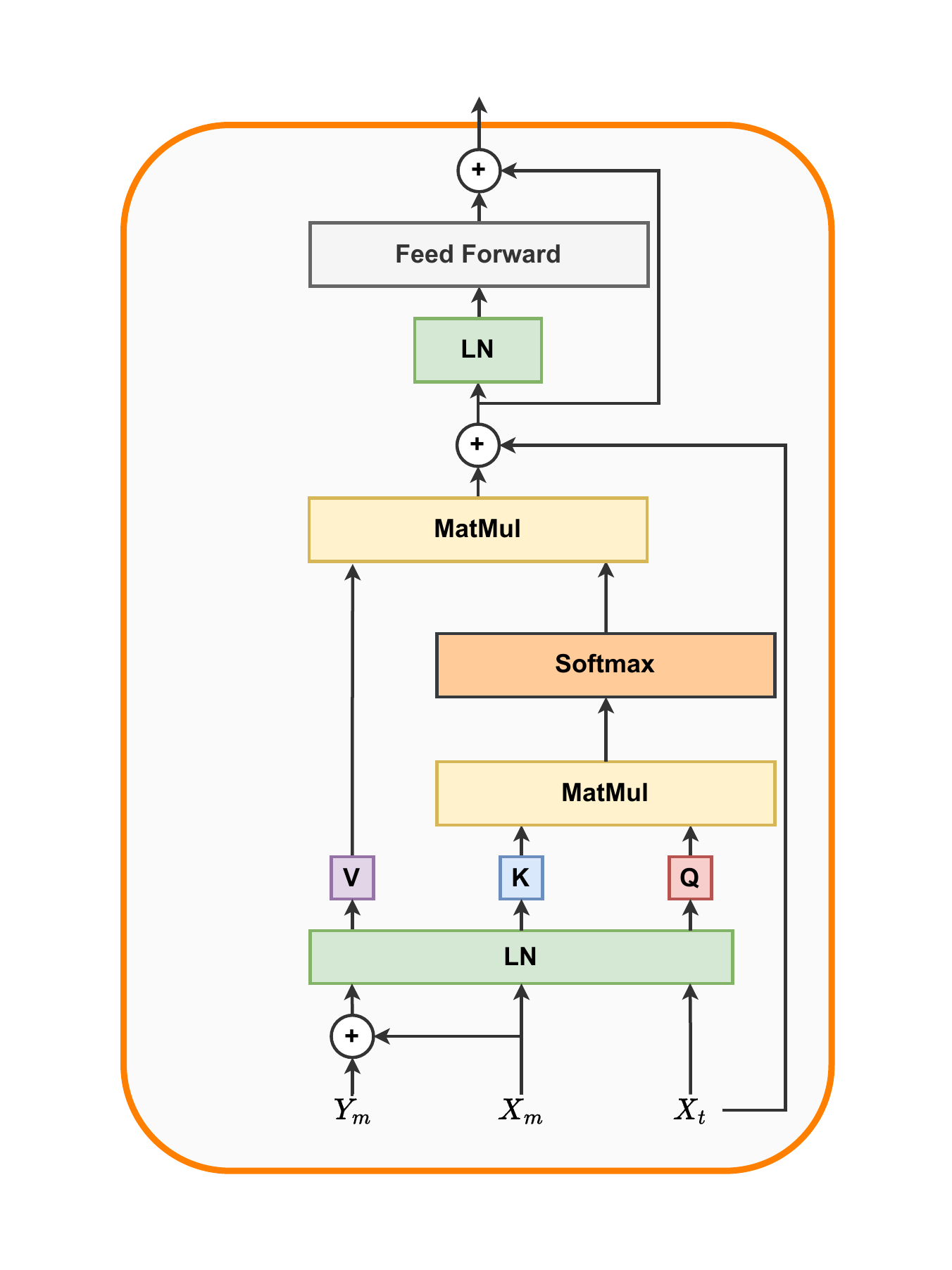}
    \caption{Cross-Attention}
    \label{fig:long}
    \end{subfigure}
\end{figure*}

The straightforward approach for VOS uses optical flow to propagate masks from the previous frame to the current frame. However, occlusions and optical flow imperfections  can lead to errors in mask propagation, degrading the quality of the propagated mask with each frame. Additionally, this approach cannot handle newly appeared parts of an object. Our proposed method, WarpFormer, aims to refine the estimated mask using semantic information, which is easier to interpret after motion was decoupled. The overall architecture of WarpFormer is shown in \Cref{fig:warpformer}.

To achieve this, some previous frame $I_k$ and mask $M_k$ is used as a reference point. Our method calculates optical flow using a given Flow Estimator:
$$f_{k\to t} = \mbox{FE}(I_k, I_t)$$
To estimate the current frame mask, the following equation is used: 
$$M_k^{warp} = \mbox{Warp}(M_{k}, f_{k\to t})$$
The method then warps the previous frame $I_k$ using the same optical flow $f_{k\to t}$ to obtain $I_k^{warp}$. Next, the features $X_t$ and $X_k$ are extracted from the current frame $I_t$ and $I_k^{warp}$ using a Feature Encoder and embedding $Y_k$ of our mask $M_k^{warp}$ is formed from an identity bank. Similarly, we create features $X_m$ and identification embedding $Y_m$ from the long-term memory frames $I_m$ with masks $M_m$. The resulting information is fed into our Refinement Transformer Block, which outputs the refined mask $\widehat{M_t}$. Finally, the decoder upsamples the refined mask estimation to the spatial dimensions of the current frame.

\subsection{Refinemenet Transformer Block}

\begin{table*}[ht!]
\caption{\textbf{The quantitative evaluation on multi-object benchmarks YouTube-VOS 2019 and DAVIS 2017.} * denotes training on MOSE 2023. Bold denotes the best or three best results. FPS in brackets denotes the value measured not including optical flow estimation time. }
\label{tab:main}
\centering
  \begin{tabular}{l|ccccc|ccc|ccc|c}
    \toprule[2pt]
    & \multicolumn{5}{c|}{YouTube-VOS 2019 Val} & \multicolumn{3}{c|}{DAVIS 2017 Val} & \multicolumn{3}{c|}{DAVIS 2017 Test}  \\
    \midrule
    Methods & $\mathcal{J}_s$ & $\mathcal{F}_s$ & $\mathcal{J}_u$ & $\mathcal{F}_u$ & $\mathcal{J}\&\mathcal{F}$ & $\mathcal{J}$ & $\mathcal{F}$ & $\mathcal{J}\&\mathcal{F}$ & $\mathcal{J}$ & $\mathcal{F}$ & $\mathcal{J}\&\mathcal{F}$ & FPS \\
    \midrule[1.5pt]
    AOT-T & 79.6 & 83.8 & 73.7 & 81.8 & 79.7 & 77.4 & 82.3 & 79.9 & 68.3 & 75.7 & 72.0 & 51.4 \\
    DeAOT-T & \textbf{81.2} & \textbf{85.6} & \textbf{76.4} & \textbf{84.7} & \textbf{82.0} & \textbf{77.7} & 83.3 & 80.5 & \textbf{70.0} & \textbf{77.3} & \textbf{73.7} & \textbf{63.5} \\
    \textbf{WarpFormer-S} & 79.0 & 85.1 & 73.5 & 82.8 & 80.1 & 77.6 & 84.2 & 80.9 & 66.2 & 76.1 & 71.1 & 27.7 (37.0) \\
    \midrule
    \textbf{WarpFormer-S\textsuperscript{*}} & 79.0 & 85.3 & 73.1 & 82.5 & 80.1 & 77.8 & \textbf{84.3} & \textbf{81.0} & 65.9 & 76.1 & 71.0 & 27.7 (37.0) \\
    \midrule[1.5pt]
    CFBI+ & 81.7 & 86.2 & 77.1 & 85.2 & 82.6 & 80.1 & 85.7 & 82.9 & 74.4 & 81.6 & 78.0 & 3.4 \\
    RMNet & 74.0 & 82.2 & 80.2 & 79.9 & 77.4 & 81.0 & 86.0 & 83.5 & 71.9 & 78.1 & 75.0 & - \\
    STCN & 81.1 & 85.4 & 78.2 & 85.9 & 82.7 & 82.2 & 88.6 & 85.4 & 73.1 & 80.0 & 76.1 & 19.5\\
    XMem & \textbf{84.3} & 88.6 & \textbf{80.3} & \textbf{88.6} & \textbf{85.5} & \textbf{82.9} & \textbf{89.5} & \textbf{86.2} & \textbf{77.4} & 84.5 & 81.0 & 20.2 \\
    ISVOS & \textbf{85.2} & \textbf{89.7} & \textbf{80.7} & \textbf{88.9} & \textbf{86.1} & \textbf{83.7} & \textbf{90.5} & \textbf{87.1} & \textbf{79.3} & \textbf{86.2} & \textbf{82.8} & - \\
    Swin-B AOT-L & 84.0 & 88.8 & 78.4 & 86.7 & 84.5 & 82.4 & 88.4 & 85.4 & 77.3 & \textbf{85.1} & \textbf{81.2} & 12.1 \\
    Swin-B DeAOT-L & \textbf{85.3} & \textbf{90.2} & \textbf{80.4} & \textbf{88.6} & \textbf{86.1} & \textbf{83.1} & 89.2 & \textbf{86.2} & \textbf{78.9} & \textbf{86.7} & \textbf{82.8} & 15.4 \\
    \textbf{WarpFormer-L} & 83.2 & 88.9 & 78.1 & 84.9 & 83.8 & 81.1 & 88.9 & 85.0 & 76.4 & 84.9 & 80.6 & 10.0 (\textbf{23.9}) \\
    \midrule
    \textbf{WarpFormer-L\textsuperscript{*}} & 83.3 & \textbf{89.1} & 78.0 & 85.0 & 83.8 & 82.4 & \textbf{89.3} & 85.9 & 76.3 & 84.9 & 80.6 & 10.0 (\textbf{23.9}) \\
    \bottomrule[1.5pt]
\end{tabular}
\end{table*}

Many recent cutting-edge VOS methods have utilized the attention mechanism and have demonstrated promising results. To define the attention mechanism formally, we consider queries ($Q$), keys ($K$), and values ($V$). The attention operation can then be defined as follows:
$$\mbox{Att}(Q, K, V) = \mbox{Corr}(Q, K)V = \operatorname{softmax}\Bigl(\frac{QK^T}{\sqrt{C}}\Bigr)V$$
where $C$ is the number of channels.

In our method, we incorporate the identification embedding into the attention operation for mask refinement as follows:
$$\mbox{AttID}(Q, K, V, ID) = \mbox{Att}(Q, K, V + ID)$$

Following the common transformer blocks, our Refinement Transformer Block (RTB) first employs a self-attention layer on the features of the images to learn the association between the targets within our frames (\Cref{fig:refinement}). Our RTB, similarly to the AOT\cite{yang2021associating}, is then divided into two branches: long-term and short-term.

The long-term branch (\Cref{fig:long}) is responsible for aggregating information from long-term (reference) memory frames. It utilizes simple cross-attention, defined as:
$$\mbox{CAtt}(X_t, X_m, Y_m) = \mbox{AttID}(X_tW_k, X_mW_k, X_mW_v, Y_m),$$
where $X_m$ and $Y_m$ are the features and masks embeddings of the long-term memory frames. Besides, $W_k$ and $W_v$ are trainable projections for matching and refinement, respectively.

The short-term (sensory memory) branch (\Cref{fig:short}) propagates information from the previous frames by taking a look at only some neighboring patches to apply matching. Since image changes between consecutive frames are smooth and continuous, this approach is only more powerful as we convert our previous frames to the current frame domain after warp. The short-term branch utilizes windowed cross-attention:
$$\mbox{WCAtt}(X_t, X_l, Y_l|p) = \mbox{Catt}(X_t^p, X_l^{N(p)}, Y_l^{N(p)})$$
where $X_l$ and $Y_l$ are the features and masks embeddings of warped previous frames, $X_t^p$ - feature of $X_t$ at location $p$ and $N(p)$ is a $\lambda \times \lambda$ spatial neighborhood centered at location $p$, where $\lambda$ is window size. We implement windowed cross-attention by including a relative position bias $B$:
$$\mbox{WCAtt}(Q, K, V) =  \operatorname{softmax}\Bigl(\frac{QK^T}{\sqrt{C}} + B\Bigr)V$$

Finally, the outputs of the long-term and short-term branches are combined together in one more self-attention layer.

\section{Implementation Details}

\subsection{Network details}

To study performance capabilities and contributions impact we introduce two variants of network architecture. Namely, \textbf{WarpFormer-S} (Small) is an efficient implementation of the proposed method, which adopts MobileNet-V2 \cite{sandler2019mobilenetv2} as encoder backbone, only a single reference frame is exploited for long-term memory. Alternatively, \textbf{WarpFormer-L} (Large) is a large-scale implementation, for which we adopt cutting edge transformer-based encoder Swin-B \cite{liu2021Swin}; following \cite{yang2021associating}, we append every \nth{2} frame to long-term memory bank for training and every \nth{5} frame for evaluation. For both architecture variants we use FPN decoder with Group Normalization \cite{lin2017feature}. We employ Global Motion Aggregation (GMA) \cite{jiang2021learning} as an optical flow estimating network for both WarpFormer-S and WarpFormer-L; however, we set the number of flow optimization updates to 4 for small architecture and to 12 for a large. 

Following \cite{yang2021associating}, we set the number of identification vectors $M$ to 10 in order to align it with the maximum object number in most of benchmarks. For encoders and patch-wise identity bank, their final resolution is $\frac{1}{16}$ as of an input image and mask. For self-attention and cross-attention blocks in Warp Refinement Transformer we use traditional multi-head architecture \cite{vaswani2017attention} with Feed-Forward layer and Layer Normalization. The embedding dimension is set to 256, the number of heads is 8 and the hidden dimension of Feed-Forward layers is 1024. For windowed cross-attention used to refine warped sensory memory, we employ original implementation \cite{liu2021Swin} with relative position bias and additionally equip learned relative positional embedding \cite{shaw2018selfattention}. The window size is set to 15. We also apply fixed sine spatial positional embedding to the self-attention following \cite{carion2020endtoend}.

\subsection{Training details}

We train both architecture variants in two stages. On the first stage, the model is trained for 40K optimization steps, while the second stage takes 60K steps. During the entire training process, we employ a mixture of DAVIS 2017 \cite{Pont-Tuset_arXiv_2017, caelles2017one} train and YouTube-VOS 2019 \cite{xu2018youtubevos1, xu2018youtubevos2} train datasets in $5:1$ proportion. Additionally, we study adopting MOSE 2023 \cite{MOSE} as additional training data, in which case we apply DAVIS, YouTube-VOS and MOSE mixture with proportion $5:k:p$ where $k+p=1$. Initial value of $k_{start}=0.5$ linearly decays during the training to a final value $k_{end}=0.25$. More detailed description of datasets is presented in \cref{sec:dataset}. For both stages we use curriculum sampling strategy \cite{oh2019video}. Notably, ground truth memory masks are used for temporal-spatial matching during the first stage, while second stage only implies an utilization of the first reference mask providing better supervision for inference setup. Identity banks are frozen after the first stage following \cite{yang2021associating}. 

We adopt AdamW optimizer \cite{loshchilov2019decoupled} with a one-cycle learning rate schedule. Initial learning rate of $lr_{start} = 3 \times 10^{-4}$ declines to a final value of $lr_{end} = 2 \times 10^{-5}$ in polynomial manner with $0.9$ decay factor. We also use learning rate warm-up \cite{gotmare2018closer} for 3000 steps. In order to prevent overfitting, we set the learning rate for the encoder to $0.1$ of the overall learning rate. Following \cite{cheng2022xmem}, we use bootstrapped cross entropy and dice losses with equal weighting. For both stages, we use a batch size of 8. WarpFormer-L model training is distributed across four RTX 3090 GPUs, while for WarpFromer-S we use only two RTX 3090 GPUs. The entire training process takes around 40 hours for the large model and 35 hours for the small one.

\subsection{Video augmentations}

\begin{figure*}[ht]
    \centering
  \includegraphics[width=\linewidth]{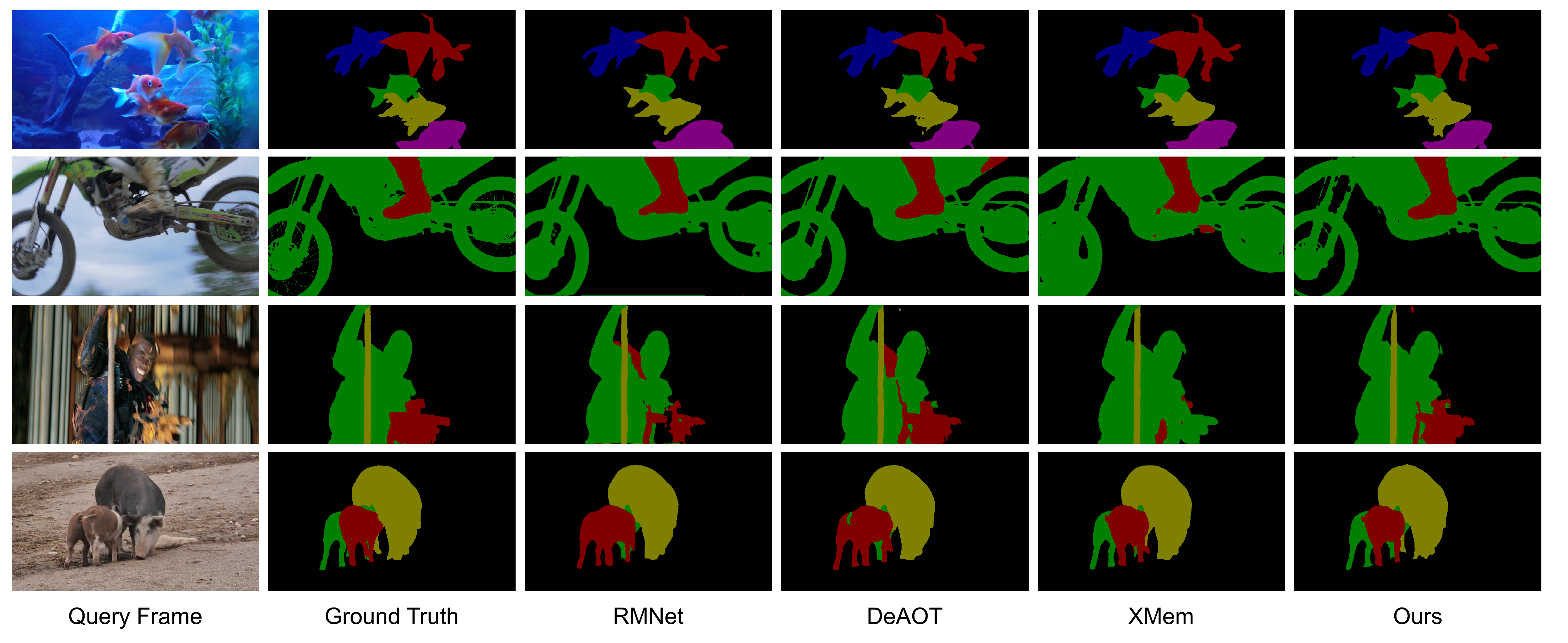}
  \caption{\textbf{Qualitative comparison between WarpFormer and several state-of-the-art VOS methods.} Best viewed in zoom. We don't include ISVOS \cite{wang2022look} since there is no source code available. For all methods we used DAVIS2017 val sequences in 480p.}
  \label{fig:qualitative}
\end{figure*}

We employ a variety of video augmentations to prevent overfitting on the seen data. Specifically, we apply random scaling followed by object-balanced random cropping to the sampled sequence. Additionally, color jitter, random Gaussian blur and random grey-scaling are applied to RGB images. 

\textbf{Dynamic merge augmentation.} In order to better adapt our model to a multi-object scenario, we adopt dynamic merging augmentation. To enrich generated sequence with more objects, we generate another sequence of the same length from a different video clip and overlay it on the top of the first one. In details, the merging process is as follows: for pair of corresponding frames from the first and second sequence the resulting frame at pixel $(x,y)$, denoted by $I_{merge}(x, y)$, is set to $I_1(x, y)$ if no objects from the second image are present at that pixel, and $I_2(x, y)$ otherwise.

For both training stages we employ the full set of augmentations, for the DAVIS and YouTube-VOS dynamic merge augmentation is applied with probability 0.4, for MOSE merge augmentation is not used since it already features complex multi-object scenes.

\section{Results}

\begin{table}[t]
\caption{\textbf{The quantitative evaluation on DAVIS 2016.} Bold denotes the best result.}
\label{tab:davis2016}
\centering
  \begin{tabular}{l|ccc}
    \toprule[2pt]
    Methods & $\mathcal{J}$ & $\mathcal{F}$ & $\mathcal{J}\&\mathcal{F}$ \\
    \midrule[1.5pt]
    AOT-T & 86.1 & 87.4 & 86.8 \\
    DeAOT-T & \textbf{87.8} & 89.9 & \textbf{88.9} \\
    \textbf{WarpFormer-S} & 87.2 & \textbf{90.5} & \textbf{88.9} \\
    \midrule[1.5pt]
    RMNet & 88.9 & 88.7 & 88.8 \\
    STCN & 90.8 & 92.5 & 91.6 \\
    XMem & 90.4 & 92.7 & 91.5 \\
    ISVOS & \textbf{91.5} & 93.7 & 92.6 \\
    Swin-B AOT-L & 90.7 & 93.3 & 92.0 \\
    Swin-B DeAOT-L & 91.1 & 94.7 & 92.9 \\
    \textbf{WarpFormer-L} & 90.7 & \textbf{95.3} & \textbf{93.0} \\
    \bottomrule[1.5pt]
\end{tabular}
\end{table}

\subsection{Metrics and Dataset}
\label{sec:dataset}

In order to evaluate our models we use traditional VOS metrics as proposed in \cite{Pont-Tuset_arXiv_2017}.

\textbf{$\boldsymbol{\mathcal{J}}$ score for region similarity evaluation.} 
$\mathcal{J}$ score (Jaccard index) is defined as the intersection-over-union (IoU) rate of the predicted and ground-truth segmentation mask. Given a predicted mask $\widehat{M}$ and ground-truth mask $G$:
$$\mathcal{J} = \frac{|\widehat{M}\displaystyle \cap G|}{|\widehat{M}\displaystyle \cup G|}$$

\textbf{$\boldsymbol{\mathcal{F}}$ score for contour accuracy evaluation.}  To estimate contour matching accuracy, one finds the contour-based precision $P_c$ and recall $R_c$ between the boundaries of the predicted and ground-truth mask. Subsequently, one computes a F1-score as a simple harmonic mean:
$$\mathcal{F} = \frac{2P_cR_c}{P_c+R_c}$$

Scores are averaged on whole video clip separately for each object. $\boldsymbol{\mathcal{J}}\boldsymbol{\&}\boldsymbol{\mathcal{F}}$ \textbf{score} is the average of $\mathcal{J}$ score and $\mathcal{F}$ score presenting a good trade-off between boundary quality and region matching.

\textbf{DAVIS 2016} \cite{Perazzi2016} is a single-object VOS benchmark  containing 20 video sequences. Even though single-object scenario is significantly less complex then the multi-object setup, the benchmark features various challenging scenarios including heavy occlusions, objects changing in shape, scale and appearance, fast movements and unfavorable environment settings.

\textbf{DAVIS 2017} \cite{Pont-Tuset_arXiv_2017} benchmark complements DAVIS 2016 with multi-object video clips. It contains 205 different objects and features a 16.1\% disappearance rate \cite{MOSE}. Benchmark presents train, validation and test-dev splits containing 60, 30 and 30 sequences respectively. While validation split doesn't introduce a high amount of unseen during training classes, test-dev is much more challenging featuring complex circumstances in most of videos. 

We evaluate our method on DAVIS 2016 \& 2017 using the default 480p 24FPS videos, not benefiting from full-resolution details. Also we do not apply any test-time augmentations like multi-scale inference \cite{chandra2016fast}.

\textbf{YouTube-VOS} \cite{xu2018youtubevos1, xu2018youtubevos2} benchmark introduces a large-scale VOS dataset covering a wide variety of in-the-wild videos. YouTube-VOS 2019 training and validation splits contain 3471, 474 video sequences respectively. Dataset features 91 object categories (7755 objects in total), 26 of which are not present in training split. The explicit annotation of unseen classes is available and the official evaluation tool additionally computes separate metrics for seen and unseen classes to benchmark the generalization power of the approaches. The disappearance rate is only 13\% \cite{MOSE}, so, in general, YouTube-VOS implies less challenging circumstances compared to DAVIS.

While evaluation our method on YouTube-VOS 2019 validation split we exploit all intermediate frames of the videos to benefit from smooth motion implying more accurate optical flow. Even though we use 24 FPS sequences during evaluation, 6FPS version is used during training and for metric computation.

\begin{table}[t]
\caption{\textbf{The quantitative evaluation on MOSE 2023.} * denotes training on MOSE 2023. Bold denotes the best result.}
\label{tab:mose2023}
\centering
  \begin{tabular}{l|ccc}
    \toprule[2pt]
    Methods & $\mathcal{J}$ & $\mathcal{F}$ & $\mathcal{J}\&\mathcal{F}$ \\
    \midrule[1.5pt]
    STCN & 46.6 & 55.0 & 50.8 \\
    RDE & 44.6 & 52.9 & 48.8 \\
    SWEM & 46.8 & 54.9 & 50.9 \\
    \textbf{WarpFormer-S\textsuperscript{*}} & \textbf{47.7} & \textbf{55.6} & \textbf{51.7} \\
    \midrule[1.5pt]
    XMem & 53.3 & 62.0 & 57.6 \\
    Swin-B AOT-L & 53.1 & 61.3 & 57.2 \\
    Swin-B DeAOT-L & 55.1 & 63.8 & 59.4 \\
    \textbf{WarpFormer-L\textsuperscript{*}} & \textbf{55.1} & \textbf{64.9} & \textbf{60.0} \\
    \bottomrule[1.5pt]
\end{tabular}
\end{table}

\textbf{MOSE 2023} \cite{MOSE} (Co\textbf{M}plex video \textbf{O}bject \textbf{SE}gmentation) is a novel VOS benchmark featuring extreme scenarios of the video sequence which are not handled good enough by existing VOS methods. The main features of introduced videos include: large number of crowded and similar objects, heavy occlusions by similar looking objects, extremely small-scale objects and reference masks covering only a small region of the whole object. MOSE contains 1507 training and 311 validation video clips with 36 object categories (5200 objects in total). MOSE features overall disappearance rate of 28.8\% which is significantly higher compared to classic VOS benchmarks. 

\subsection{Comparison with State-of-the-art Methods}

Our method doesn't adopt complex memory model used in existing methods (XMem \cite{cheng2022xmem}), neither it features special architecture injecting instance segmentation logic to benefit from better instance-specific understanding (ISVOS \cite{wang2022look}). Also both our small and large models feature only a single transformer block for spatial-temporal matching while existing methods (AOT \cite{yang2021associating}, DeAOT \cite{yang2021decoupling}) use up to three blocks. Instead, we incorporate additional training data from MOSE 2023, allowing WarpFormer to tackle scenarios with heavy occlusions, large number of overlapping similar objects or objects dramatically changing in appearance and scale. 

\textbf{Quantitative comparison.} The comparison of WarpFormer with other state-of-the-art methods on DAVIS 2017 validation, DAVIS 2017 test-dev and Youtube-VOS 2019 validation validation may be found in \Cref{tab:main}. The quantitative comparison with relevant existing methods on DAVIS 2016 validation are listed in \Cref{tab:davis2016}.

Without training on MOSE 2023, our Swin-B WarpFormer-L achieves state-of-the-art performance on DAVIS 2016 single-object benchmark scoring \textbf{93.0\%} $\mathcal{J}\&\mathcal{F}$. Being evaluated on multi-object benchmarks, model demonstrates highly competitive performance wrapping up with top-ranked scores \ie \textbf{85.0\%} and \textbf{80.6\%} $\mathcal{J}\&\mathcal{F}$ on DAVIS 2017 validation and test-dev splits and \textbf{83.8\%} $\mathcal{J}\&\mathcal{F}$ on Youtube-VOS 2019 validation.

Trained only on Youtube-VOS and DAVIS, our MobileNet-V2 WarpFormer-S outperforms most of its competitors on both single-object and multi-object benchmarks. Namely, it scores \textbf{88.9\%}, \textbf{81.0\%} and \textbf{71.0\%} $\mathcal{J}\&\mathcal{F}$ on DAVIS 2016 validation and DAVIS 2017 validation \& test-dev. YouTube-VOS 2019 validation score is \textbf{80.1\%} $\mathcal{J}\&\mathcal{F}$. We believe that strong and balanced performance under different complex scenarios, simple architecture and lightweight encoder along with agnosticity of actual flow estimation method make WarpFormer-S ideal candidate for usage in various industrial applications.

\textbf{Qualitative comparison.} The qualitative comparison of state-of-the-art approaches and our method is visualized in \cref{fig:qualitative}. Existing methods fail to reconstruct fine-grained details under the rapid motion circumstances. In contrast, our method benefits from global motion field and is much more robust to motion blur. On the other hand, adopting MOSE as additional training data gives enough supervision to successfully handle overlapping similar objects without having special architecture design, as instance segmentation branch \cite{wang2022look} or feature decoupling module \cite{yang2021decoupling}.

\subsection{Training with MOSE 2023}

Adopting MOSE 2023 as training data gives a significant boost on MOSE 2023 validation split so that both our WarpFormer-S and WarpFormer-L models achieve state-of-the-art performance among competitors, scoring \textbf{51.7\%} and \textbf{60.0\%} $\mathcal{J}\&\mathcal{F}$ respectively. One the other hand, performance on the classic benchmarks experience an insignificant boost, likely because they doesn't feature any similar extreme scenarios. However, they focus on circumstances with a large number of object classes and classes unseen during training, along with a wide variety of challenging environments, while MOSE 2023 lacks such flexibility. Wrapping up, even minor improvements on classic benchmarks while training with MOSE 2023 indicate the high robustness and performance capacity of the proposed method. The quantitative comparison with other methods on MOSE 2023 validation are listed in \Cref{tab:mose2023}.

\subsection{Optical Flow benchmark}

We benchmark different optical flow estimation methods during evaluation on DAVIS 2017. As our architecture is completely agnostic to the actual implementation of the flow estimator, we test various approaches in terms of performance / resource requirements trade-off. For RAFT-based models \cite{teed2020raft, jiang2021learning, huang2022flowformer}, we also try various numbers of iterative flow updates. To demonstrate the impact of flow-warped windowed attention refinement, we also include "zero-flow", which implies identity transformation; in this case, our sensory memory processing degenerates to simple windowed attention similar to \cite{yang2021associating}. The quantitative comparison may be found in \Cref{tab:flow}.

The results indicate that our model is indeed optical flow agnostic, and its performance is directly proportional to the quality of the flow. Additionally, for iterative-based optical flow approaches, we observed that a smaller number of iterations was sufficient to achieve fairly good results. This may be attributed to the model's ability to already capture the global motion trend. However, the accuracy of "zero-flow" deteriorated, as our network was trained solely for refinement, rather than direct matching.

\begin{table}[ht!]
\caption{\textbf{Optical Flow estimator benchmark.} Subscript denotes the number of flow optimization iterations.}
\label{tab:flow}
\centering
  \begin{tabular}{l|c|cc}
    \toprule[2pt]
    Methods & $\mathcal{J}\&\mathcal{F}$ & $\#\text{param.}$ & FPS\\
    \midrule
    \multicolumn{4}{c}{MobileNet-V2} \\
    \midrule
    Zero-Flow & 76.1 & 7.7M & 57.8 \\
    RAFT-S\textsubscript{4} & 80.5 & 8.7M & 34.7 \\
    RAFT\textsubscript{4} & 80.7 & 13M & 33.6 \\
    RAFT\textsubscript{12} & 80.7 & 13M & 18.4 \\
    GMA\textsubscript{1} & 80.2 & 13.6M & 37.0 \\
    GMA\textsubscript{4} & 80.8 & 13.6M & 27.7 \\
    GMA\textsubscript{12} & 81.0 & 13.6M & 12.6 \\
    GMA\textsubscript{32} & 80.8 & 13.6M & 6.1 \\
    FlowFormer & 80.7 & 23.9M & 3.9 \\
    \midrule
    \multicolumn{4}{c}{Swin-B} \\
    \midrule
    Zero-Flow & 80.7 & 64.9M & 32.2 \\
    GMA\textsubscript{1} & 85.0 & 70.8M & 23.9 \\
    GMA\textsubscript{4} & 85.7 & 70.8M & 15.2 \\
    GMA\textsubscript{12} & 85.9 & 70.8M & 10.0 \\
    FlowFormer & 85.9 & 81.1M & 3.6 \\
    \bottomrule[1.5pt]
\end{tabular}
\end{table}

\section{Conclusion}

This paper proposes to reuse existing motion understanding knowledge by adopting optical flow estimation network to support a generic VOS architecture. To integrate global motion structure we replace propagation with optical flow warping and introduce Warp Refinement Transformer block, which aims to inpaint occlusions and fuse warped segmentation mask with long-term memory information. Experimental results show that our method demonstrates strong performance and generalization capabilities. We believe that combining WarpFormer with complex memory mechanisms or specific architecture blocks for instance understanding may further boost it effectivness.

\section{Acknowledgements}

The work is supported by Ukrainian Catholic University and ADVA Soft.

%%%%%%%%% REFERENCES
{\small
\bibliographystyle{ieee_fullname}
\bibliography{egbib}
}

\end{document}